# Finding frames with BERT: A transformer-based approach to generic news frame detection

Vihang Jumle*, Mykola Makhortykh*, Maryna Sydorova*, and Victoria Vziatysheva*

*Institute of Communication and Media Studies, University of Bern

*Corresponding author*

Vihang Jumle, Institute of Communication and Media Studies, University of Bern, Fabrikstrasse 8, 3012 Bern, phone: 031 631 48 11, e-mail: vihang.jumle@unibe.ch

*Declaration of conflicting interest*

No potential conflict of interest was reported by the author(s).

*Funding statement*

This article is part of the project "Algorithm audit of the impact of user- and system-side factors on web search bias in the context of federal popular votes in Switzerland" (PI: Mykola Makhortykh) funded by the Swiss National Science Foundation [105217_215021].

*Ethical approval and informed consent statements*

The design of the project of which the article is part has been approved by the Ethics Committee of the Faculty of Business, Economics and Social Sciences of the University of Bern (serial number 382023).

# Finding frames with BERT: A transformer-based approach to generic news frame detection

**Abstract**: Framing is among the most extensively used concepts in the field of communication science. The availability of digital data offers new possibilities for studying how specific aspects of social reality are made more salient in online communication but also raises challenges related to the scaling of framing analysis and its adoption to new research areas (e.g. studying the impact of artificial intelligence-powered systems on representation of societally relevant issues). To address these challenges, we introduce a transformer-based approach for generic news frame detection in Anglophone online content. While doing so, we discuss the composition of the training and test datasets, the model architecture, and the validation of the approach and reflect on the possibilities and limitations of the automated detection of generic news frames.

## Introduction

Defined by Entmann (1993) as a process of selecting and making more salient the specific aspects of social reality, framing is among the most extensively used concepts in the field of communication science (Olsson & Ihlen, 2018). The abundant body of research utilising the concept of framing highlights the versatility of the concept: it has been used for examining the representation of armed conflict (Tschirky & Makhortykh, 2024), climate change (Vu et al., 2021), politics (Ogan et al., 2018), and racial injustice (Lane et al., 2020). The diversity of areas in which the concept of framing is applied and the vagueness of its operationalisation are, however, occasionally viewed as the concept's weakness: Cacciatore et al. (2016) note that it results in the unnecessarily broad understanding of framing that overlaps with other concepts, such as agenda-setting, and diminishes its explanatory potential.

Despite the above-mentioned criticism, we suggest that framing remains an essential tool for understanding how certain interpretations of important societal issues become more visible and in which ways individuals are exposed to these interpretations. The importance of such an understanding increases under the conditions of the high-choice media environment (van Aelst et al., 2017) in which we are consuming information. With more available information sources and, consequently, more possibilities for being exposed to them — both selectively (Messing & Westwood, 2014) and incidentally (Lee & Kim, 2014) — it is crucial to be able to distinguish between frames coming from these sources, especially regarding the salience of epistemically contested issues which can easily amplify polarisation in the society. The ability to detect the presence or absence of specific frames in this context also becomes paramount for detecting attempts to manipulate public opinion.

Another reason why frame detection is highly relevant is the growing reliance on artificial intelligence (AI)-powered systems for organising and generating information regarding societally relevant issues. The adoption of systems such as search engines and recommendations systems and, recently, generative AI-powered chatbots has profound implications for how individuals are exposed to information as these systems decide what information sources and interpretations are prioritised in response to the user input (e.g. search queries or chatbot prompts). Under the conditions of a high-choice media environment, the use of such systems is almost inevitable due to the overabundance of available information; however, their adoption also raises concerns about the differentiated treatment of individual users that can contribute to the information inequalities between population groups.

In this short note, we discuss how we can implement large-scale frame analysis using natural language processing for facilitating different forms of communication research (e.g. focused on information exposure). Specifically, we focus on the detection of generic news frames, which is the category of frames introduced by Semetko and Valkenburg (2000). Generic frames are those frames that tend to be "persistent over time" (Reese, 2001), namely, human interest, conflict, morality, economic frame, and attribution of responsibility. Unlike issue-specific frames, which can vary broadly depending on the subject being represented, generic news frames are related to general journalistic conventions of representing diverse subjects (de Vreese, 2005). It makes generic news frames more applicable for comparing how different issues are framed and such comparability of particular relevance for studying unequal exposure to specific interpretations. In our specific use case, we are particularly interested in the exposure to generic news frames via outputs of search engines, but the approach we propose can also be easily adapted to other contexts.

The rest of the paper is organised as follows: first, we provide a short overview of the related work on the automated detection of the (news) frames. This is followed by a description of the transformer-based approach that we propose, including information about training data, model architecture and hyperparameters, and the results of the model validation. We conclude with a short summary of our approach together with a discussion of the possible use cases and the limitations of the current work.

**Related work**

For a long time framing analyses — either qualitative or quantitative — dealt with a small amount of data. Manual labelling was the central component of these approaches, and each unit of analysis would be studied by the researcher to ascertain the presence of a particular frame. However, the ease of access to the growing volumes of digital data prompted the adoption of computational methods for framing (Card et al., 2015; Liu et al., 2019; Walter and Ophir, 2019). This shift came with a set of advantages. Firstly, it allowed scaling the framing analyses by accelerating the analysis of (predominantly textual) data and decreasing the amount of required resources. Secondly, the computational approaches showed the capacity to be (at least partially) consistent with the idea of frame detection.

Most commonly employed methods for frame detection — i.e. topic modelling ( including its structural and hierarchical versions), clustering and semantic analyses or combinations of these methods, produced consistently topical or thematic insights. However, whether the outputs of these methods are sufficient to qualify for framing analysis remains inconclusive (Ali & Hassan, 2022). At most, these methods provide a broad thematic breakdown (sometimes too broad) of the content included in the dataset that may be insufficient to qualify as frame detection since it does not capture linguistic and semantic nuances to the extent conceptualised by Entman (1993) who noted that frames explain not only "what" words are present, but "how" are they presented. These shortcomings are particularly evident since outputs are primarily generated based on word frequency or collections (like bag-of-words that do not capture the relationship between these word collections) or their corpus distribution and given that these methods were not specifically developed for frame detection in the first place. Such applications are, therefore, methodological innovations that are prone to their own set of limitations and sometimes demand a refitting of framing in itself to the method in use.

What, however, seems to be addressing these shortcomings is the use of neural network models. Particularly promising is the application of transformer models (for instance, BERT) for automated frame detection using transformers that often outperform existing methods for computational text analysis, like support vector machines (Khanehzar et al., 2019). Ali and Hassan (2022), in their review of the field, found likewise. These reviewed studies used different computational methods (e.g. long short-term memory networks and transformers) predominantly to analyse two annotated datasets: the media frames corpus (MFC; Card et al., 2015) and the gun violence frame corpus (GVFC; Liu et al., 2019). The comparative review showed that these neural network-based methods were "promising for advancing computational framing analysis" and that the models presented good accuracy. For instance, Liu et al. (2019) used BERT to automatically detect a diverse set of generic frames and achieved an accuracy of 0.84[1]. Others like Akyürek et al. (2020) and Tourni et al. (2021) have similarly explored the GVFC dataset using BERT towards multilingual frame detection and improving generic frame detection by complimenting news headlines with images, respectively.

Ali and Hassan (2022), however, also identified shortcomings in the existing training datasets, calling them "not up to the mark" and limiting the potential of frame detection approaches using them. Implied in this criticism was that datasets often do not capture adequate framing nuance. For instance, the MFC dataset made use of the label "economic frame" (derived from the Policy Frames codebook by Boydstun et al. (2013)) when a news report focused on "the costs, benefits, or monetary/financial implications of the issue (to an individual, family, community, or to the economy as a whole)". Ali and Hassan (2022) say that it is appropriate to term these dimensions as topics rather than frames. Liu et al. (2019), who used the GVFC dataset, manoeuvred this criticism by focusing on word choices and increasing the conceptual proximity to Entman's (1993) framing definition. Ali and Hassan (2022), however, note that the criteria to identify the "politics" frame in GVFC was still defined too broadly as against the nuances that a frame would ideally demand.

---

[1] They propose a 5-step process to conduct computational framing analysis (combing topic modelling for frame identification and BERT for automated frame detection) of news reports allowing the identification of both generic and issue-specific news frames.

Our approach in this paper is similar to Liu et al. (2019), although with a few modifications and advancements. First, we do not construct topic models and focus solely on generic frame detection. Second, we build our own labelled data corpus that includes all kinds of information sources (in addition to news articles). Third, while Liu et al. (2019) demonstrate their model efficacy based on analysing news headlines, we demonstrate it at the level of more typical forms of online content (i.e. our unit of analysis is a paragraph within a text corpus, such as a news article). Our contribution, therefore, helps situate the application of language models like BERT in generic framing analysis beyond journalistic content and on forms of textual content other than headlines. We also address the criticism raised by Ali and Hassan (2022) towards the operationalisation of frames in the method section below.

**Method**

In this section, we introduce our approach for generic frame detection. The model described below is publicly available [anonymously for peer review] at this link: https://osf.io/vwcgd/?view_only=d1f2d95ca3894cf9bc26761ed9e9c35a.

*Operationalization of generic news frames*

As noted above, this paper proposes a computational approach aiming to automatically detect generic news frames. We focus on five generic news frames proposed by Semetko and Valkenburg (2000): human interest, conflict, morality, economic frame, and attribution of responsibility. While we generally follow the conceptual framework of Semetko and Valkenburg - also in understanding frames as specific patterns of representing, conveying and interpreting information - we operationalise these frames slightly differently for our approach. While Semetko and Valkenburg (2000) used multiple questions per frame to manually assess the presence of each generic frame, we abstracted these questions into a maximum of two questions per frame for preparing data for automated detection. This simplification was made to facilitate the process of training coders for training data labelling and for focusing on more pronounced features of each frame while retaining the core meaning of frames.

The codebook used for preparing the training data consisted of six variables which we describe below. As a unit of coding, we used a paragraph because individual sentences were potentially too short to capture individual frames, and the whole document was difficult to code due to the length and likely containing multiple frames. Except for the last variable — i.e. No frame — the other five variables were not exclusive (i.e. the same paragraph could be labelled as containing several frames; however, we asked coders to mark the frame which, in their view, is the most pronounced in the paragraph). The model discussed below has been used specifically to predict the main frame, albeit the data labelling also provides possibilities for multi-label classification.

*Attribution of responsibility (AR01)*

1. Does the paragraph suggest that an individual (e.g. a politician) or a group (e.g. a party, state, governmental departments, society, civilian groups) is responsible for the problem or can resolve it?
2. Does the paragraph suggest a solution to the problem or call for urgent action over it?

*Human interest frame (HI02)*

1. Does the paragraph use a human example or emphasise the effect of a problem on humans? Is the human the central focus of the paragraph?
2. Does the paragraph use emotive language that may invoke an emotional response in the reader (like outrage, empathy-caring, sympathy, or compassion)?

*Conflict frame (CF03)*

1. Does the paragraph refer to any form of negative interaction or framing (disagreement, confrontation, spat, etc.) between two sides of any kind (actors, problem, viewpoints)?

*Morality frame (MF04)*

1. Does the paragraph contain <u>any form of morality</u> (what is good or bad? Does it talk about Good and Evil? Does it talk about religious tenets or prescribe a socially apt behaviour or ethics?)

*Economic frame (EF05)*

1. Does the paragraph speak about economic changes in policy or law or refer to <u>any form of economic loss/gain, expense, costs, or economic consequences of a current policy or law</u>?

*No frame (NO06)*

It is also important to note the limitations of the operationalisation we chose. We agree with Ali and Hassan (2022) that automated frame detection is inevitably prone to more errors and capturing less nuance than manual labelling conducted by trained researchers. This is because manual labelling involves elements of interpretation that are informed by the implicit knowledge of the involved researcher and help in identifying less obvious frames. Such interpretation is difficult to comprehensively translate into coding criteria, which can be used for automated detection and rely on preset and rigid computer logic. A common response to these hurdles, as shown by existing research, has been to expand the framing criteria substantially (i.e. by searching for the presence of a particular keyword instead of complicating the logic over its contextual use in a corpus). Our study, on the other hand, takes a lean approach towards frame detection wherein we aim to identify if certain defining features (underlined in the codes above) of a particular frame mark their presence in the data corpus rather than exploring a frame comprehensively. The "no frame" option, therefore, helps us to not force fit a frame onto a paragraph if it does not meet the defined criteria. We suggest that the computational identification of defining features can be used as an entry point for a second level of manual labelling to address the framing nuances in more detail.

*Training data*

Our training dataset consists of web pages that were collected via an algorithm audit conducted for two search engines (Google and Bing) with a set of queries collected via the representative survey of the Swiss population in January 2024. For the survey, we asked participants to provide search queries that they would use to find information about two Swiss popular votes in March 2024. The topics of the popular votes concerned retirement policies. In particular, citizens voted for two popular initiatives: (1) the Initiative for a 13th OASI pension payment, which proposed an extra payment each year for the recipients of old-age and survivors' insurance (OASI) pension, and (2) the Pensions initiative, which was calling for the retirement age for women and men to be gradually increased to 66 and then tied to the life expectancy. In the course of the virtual agent-based audit (for the discussion of the method, see Ulloa et al., 2024), we programmed virtual agents to collect the top 10 search results for Google and Bing text search. These search results were then crawled and sampled to be used as training data.

To prepare the training dataset, each crawled web page was translated into English (from German, French, or Italian, if not already in English) using Google Translation API and split into paragraphs (as they appeared in the web pages). The unit of our analysis is, therefore, a paragraph. 372 web pages were selected at random, yielding 2,736 paragraphs. These 2,736 paragraphs were manually labelled using the procedure described above. Additionally, 119 paragraphs selected from randomly selected web pages were also manually labelled to be used as a golden standard for model testing. All these paragraphs were labelled by a single research assistant.

The distribution of paragraphs according to the prevalent frame is shown in Table 1 below. Before manually labelling training and test datasets, a sample coding exercise was conducted to test the performance of the research assistant to ensure intercoder reliability. 35 paragraphs were coded independently by the research assistant and one of the authors responsible for the development of the codebook and coding instructions. Cohen's Kappa was calculated to estimate the agreement between the coders. The analysis yielded a Kappa value of $\kappa = 0.74$ (95% CI: 0.57-0.74), which is considered a substantial agreement (Landis & Koch, 1977).

**Table 1**: Distribution of manually labelled frames

| Frames (Manually Labelled) | Count |
|---|---|
| AR01 | 541 |
| HI02 | 780 |
| CF03 | 83 |
| MF04 | 14 |
| EF05 | 365 |
| NO06 | 953 |
| **Grand Total** | **2,736** |

*Model architecture and training process*

The model architecture utilises BERT (Bidirectional Encoder Representations from Transformers; Devlin, 2018), specifically the bert-base-uncased variant, tailored for sequence classification tasks and pre-trained by Hugging Face. For fine-tuning it for our task, the model was initialised with a number of labels equivalent to the described number of frames. This architecture leverages the powerful capabilities of BERT to understand and process natural language, making it well-suited for tasks requiring deep linguistic comprehension.

The training process involved additional finetuning with several hyperparameters to optimise performance. The model outputs to the directory ./results, with an overwrite option enabled to ensure the latest outputs are always saved. Batch sizes are set to 4 for both training and evaluation, and the training spans five epochs. A learning rate of $2 \times 10^{-5}$ is employed, balancing the need for effective learning while avoiding potential overfitting. Logging is configured to occur every ten steps, with logs stored in the ./logs directory and reporting enabled for TensorBoard, facilitating detailed tracking and visualisation of the training process. The Trainer class, combined with a data collator for padding, ensures efficient and streamlined model training and evaluation. The training is set up using the Trainer class from the Hugging Face Transformers library, with data collated using the DataCollatorWithPadding. The training process includes callbacks for TensorBoard logging. The model and tokenizer are both initialised from the bert-base-uncased pre-trained model.

The model undergoes regular validation. The Trainer class from the Hugging Face Transformers library periodically evaluates the model's performance at intervals specified by the logging_steps parameter against the evaluation dataset.

*Model validation*

The model projected an average macro F1 score of 0.92 based on the golden standard used to evaluate the model's performance. Other metrics are summarised in Figure 1 below.

```
+-----------+----------+--------+
|   Metric  | Weighted | Macro  |
+-----------+----------+--------+
| Precision |  0.9338  | 0.9285 |
|    Recall |  0.9328  | 0.9289 |
|  F1 Score |  0.9298  | 0.9242 |
+-----------+----------+--------+
```

**Figure 1**. Performance metrics for the BERT model used for generic news frame detection

The model was further checked for robustness using a 5-fold cross-validation with results shown in Figure 2. Overall accuracy stood at 0.98. Four classes (AR01, EF05, HI02, NO06) consistently reported high precision, recall and F1, close to 0.98. CF03 reported a slightly weaker precision (0.88), recall (0.92) and F1 (0.9). This is due to the relatively lower count of the class in the training data. MF04, for similar reasons, predicts these metrics at 0.0.

```
Classification Report:
+-----------+--------+-------+--------+--------+-------+--------+
|   Metric  |  AR01  | CF03  |  EF05  |  HI02  | MF04  |  NO06  |
+-----------+--------+-------+--------+--------+-------+--------+
| precision |  0.97  | 0.88  |  1.00  |  0.98  | 0.00  |  0.99  |
|    recall |  0.99  | 0.92  |  0.99  |  0.99  | 0.00  |  0.99  |
|  f1-score |  0.98  | 0.90  |  0.99  |  0.99  | 0.00  |  0.99  |
|   support | 541.00 | 83.00 | 365.00 | 780.00 | 14.00 | 953.00 |
+-----------+--------+-------+--------+--------+-------+--------+
```

**Figure 2**. The classification report for the 5-fold cross-validation

The weighted average and confusion matrix (Figure 3) show that the metrics largely perform well, except when predicting class MF04, which is attributed to the limited number of data points for this class in the training data.

```
Confusion Matrix:
+-------------+---------------+---------------+---------------+---------------+---------------+---------------+
|             | Predicted AR01| Predicted CF03| Predicted EF05| Predicted HI02| Predicted MF04| Predicted NO06|
+-------------+---------------+---------------+---------------+---------------+---------------+---------------+
| Actual AR01 |      534      |       2       |       0       |       3       |       0       |       2       |
| Actual CF03 |       2       |      76       |       1       |       1       |       0       |       3       |
| Actual EF05 |       0       |       0       |      361      |       2       |       0       |       2       |
| Actual HI02 |       7       |       1       |       0       |      772      |       0       |       0       |
| Actual MF04 |       2       |       6       |       0       |       3       |       0       |       3       |
| Actual NO06 |       3       |       1       |       0       |       3       |       0       |      946      |
+-------------+---------------+---------------+---------------+---------------+---------------+---------------+
```

**Figure 3**. Confusion matrix for the predicted generic news frame classes

**Conclusions**

In this paper, we introduced an automated generic news frame detection approach which is based on BERT, a popular transformer-based language model. In addition to discussing the different stages of the development of the approach, we also publicly released the model for reuse by the community. The current model is specifically adapted to identify five generic frames in English language content in the context of retirement-related topics. Despite considering the nature of generic news frames, it has the potential to be adapted for other contexts.

    In addition to the practical contribution in the form of the model release, our paper showcases the impressive potential of transformer-based approaches for automated content analysis in the fields of communication and social science. Not only do transformer-based approaches show high performance, but they also require relatively little training resources. Already, with the training dataset of approximately

1,000 paragraphs, the model showed solid performance, reaching the macro average F1 score of approximately 0.8. With the further addition of training data, it reached a macro average F1 score of 0.92. Such a performance demonstrates that with a rather small training dataset (which can further be expanded using AI-generated synthetic data), transformer-based approaches can be adapted for highly complex computational tasks and diverse cases. While in our case, we were specifically interested in using the approach for evaluating the representation of the popular votes by search engines, there are many other use cases. Some of them include, for instance, the analysis of journalistic coverage of contested topics such as elections or armed conflicts, the comparison of how different social media communities represent particular issues or even large-scale examination of audiovisual framing on platforms like YouTube or TikTok (e.g. in combination with speech-to-text models).

It is also important to note several limitations of the current study. First of all, despite the universalist nature of generic frames, the context in which they are used has implications for how they are formulated and expressed. Considering our focus on a specific context outlined earlier, it is crucial to evaluate the performance of our approach for frame detection regarding other issues and in other domains. Second, we focused on English language content: while the training data for our use case was originally in non-English languages, we decided to automatically translate it to avoid developing three separate detection models. While this approach can result in some semantic losses, we found it more feasible and resource-affordable. However, for future research, it would be beneficial to examine in detail the possible differences in the performance of models trained on data in the original language versus models relying on the translated data. Third, while achieving solid performance for detecting four generic frames out of five, the last generic frame — i.e. morality — is not detected properly due to the lack of corresponding content in the training data.

**References**

Akyürek, A. F., Guo, L., Elanwar, R., Ishwar, P., Betke, M., & Wijaya, D. T. (2020). Multi-label and multilingual news framing analysis. In *Proceedings of the 58th*

*Annual Meeting of the Association for Computational Linguistics* (pp. 8614-8624). ACL. https://doi.org/10.18653/v1/2020.acl-main.763

Ali, M., & Hassan, N. (2022). A survey of computational framing analysis approaches. In *Proceedings of the 2022 Conference on Empirical Methods in Natural Language Processing* (pp. 9335-9348). ACL. https://doi.org/10.18653/v1/2022.emnlp-main.633

Boydstun, A. E., Gross, J. H., Resnik, P., & Smith, N. A. (2013). Identifying media frames and frame dynamics within and across policy issues. In *New Directions in Analyzing Text as Data Workshop* (pp. 1-13). https://faculty.washington.edu/jwilker/559/frames-2013.pdf

Cacciatore, M. A., Scheufele, D. A., & Iyengar, S. (2016). The end of framing as we know it… and the future of media effects. *Mass Communication and Society*, *19*(1), 7-23. https://doi.org/10.1080/15205436.2015.1068811

Card, D., Boydstun, A., Gross, J. H., Resnik, P., & Smith, N. A. (2015). The media frames corpus: Annotations of frames across issues. In *Proceedings of the 53rd Annual Meeting of the Association for Computational Linguistics and the 7th International Joint Conference on Natural Language Processing* (pp. 438-444). ACL. https://doi.org/10.3115/v1/P15-2072

De Vreese, C. H. (2005). News framing: Theory and typology. *Information Design Journal + Document Design*, *13*(1), 51-62.

Devlin, J. (2018). Bert: Pre-training of deep bidirectional transformers for language understanding. arXiv. https://doi.org/10.48550/arXiv.1810.04805

Entman, R. M. (1993). Framing: Toward clarification of a fractured paradigm. *Journal of Communication*, *43*(4), 51-58. https://doi.org/10.1111/j.1460-2466.1993.tb01304.x

Khanehzar, S., Turpin, A., & Mikolajczak, G. (2019). Modeling political framing across policy issues and contexts. In *Proceedings of The 17th Annual Workshop of the Australasian Language Technology Association* (pp. 61-66). ACL.

Lane, K., Williams, Y., Hunt, A. N., & Paulk, A. (2020). The framing of race: Trayvon Martin and the Black Lives Matter movement. *Journal of Black Studies, 51*(8), 790-812. https://doi.org/10.1177/0021934720946802

Landis, J. R., & Koch, G. G. (1977). The measurement of observer agreement for categorical data. *Biometrics*, 33(1), 159–174.

Lee, J. K., & Kim, E. (2017). Incidental exposure to news: Predictors in the social media setting and effects on information gain online. *Computers in Human Behavior, 75*, 1008-1015. https://doi.org/10.1016/j.chb.2017.02.018

Liu, S., Guo, L., Mays, K., Betke, M., & Wijaya, D. T. (2019). Detecting frames in news headlines and its application to analyzing news framing trends surrounding US gun violence. In *Proceedings of the 23rd conference on computational natural language learning* (pp. 504-514). ACL. https://doi.org/10.18653/v1/K19-1047

Messing, S., & Westwood, S. J. (2014). Selective exposure in the age of social media: Endorsements trump partisan source affiliation when selecting news online. *Communication Research, 41*(8), 1042-1063. https://doi.org/10.1177/0093650212466406

Ogan, C., Pennington, R., Venger, O., & Metz, D. (2018). Who drove the discourse? News coverage and policy framing of immigrants and refugees in the 2016 US presidential election. *Communications, 43*(3), 357-378. https://doi.org/10.1515/commun-2018-0014

Olsson, E. K., & Ihlen, Ø. (2018). Framing. In R. L. Heath & W. Johansen (Eds.), *The International Encyclopedia of Strategic Communication* (pp. 1-11). https://doi.org/10.1002/9781119010722.iesc0076

Semetko, H. A., & Valkenburg, P. M. (2000). Framing European politics: A content analysis of press and television news. *Journal of Communication, 50*(2), 93-109. https://doi.org/10.1111/j.1460-2466.2000.tb02843.x

Tourni, I., Guo, L., Daryanto, T. H., Zhafransyah, F., Halim, E. E., Jalal, M., ... & Wijaya, D. T. (2021). Detecting frames in news headlines and lead images in US gun violence coverage. In *Findings of the Association for Computational Linguistics: EMNLP 2021* (pp. 4037-4050). ACL. https://doi.org/10.18653/v1/2021.findings-emnlp.339

Tschirky, M., & Makhortykh, M. (2024). #Azovsteel: Comparing qualitative and quantitative approaches for studying framing of the siege of Mariupol on Twitter. *Media, War & Conflict*, 17(2), 163-178. https://doi.org/10.1177/17506352231184163

Van Aelst, P., Strömbäck, J., Aalberg, T., Esser, F., De Vreese, C., Matthes, J., ... & Stanyer, J. (2017). Political communication in a high-choice media environment: a challenge for democracy? *Annals of the International Communication Association, 41*(1), 3-27. https://doi.org/10.1080/23808985.2017.1288551

Vu, H. T., Blomberg, M., Seo, H., Liu, Y., Shayesteh, F., & Do, H. V. (2021). Social media and environmental activism: Framing climate change on Facebook by global NGOs. *Science Communication*, 43(1), 91-115. https://doi.org/10.1177/1075547020971644

Walter, D., & Ophir, Y. (2019). News frame analysis: An inductive mixed-method computational approach. *Communication Methods and Measures*, *13*(4), 248-266. https://doi.org/10.1080/19312458.2019.1639145